\documentclass[9pt, journal, twocolumn]{IEEEtran}

\usepackage{graphicx} % for including graphics
\usepackage{amsmath} % for advanced math formatting
\usepackage{cite} % for IEEE style citation
\usepackage{caption} % Improved caption handling
\usepackage{threeparttable} % For adding notes below the table
\usepackage{booktabs} % For prettier tables
\usepackage{xcolor}
\usepackage[colorlinks=true,linkcolor=blue,citecolor=blue,urlcolor=blue]{hyperref}
\usepackage{helvet} % Arial font
\usepackage{sansmath} % Sans-serif math

\usepackage{scalerel}
\usepackage{graphicx}

\newcommand{\orcidicon}{
    \scalerel*{
        \includegraphics{orcid.pdf}
    }{A}
}

\newcommand\orcid[1]{\href{https://orcid.org/#1}{\orcidicon}}

\title{Redefining Aerial Innovation: Autonomous Tethered Drones as a Solution to Battery Life and Data Latency Challenges}
\title{Redefining Aerial Innovation: Autonomous Tethered Drones as a Solution to Battery Life and Data Latency Challenges}

\author{Samuel O. Folorunsho \orcid{0000-0001-6386-9190}, \IEEEmembership{Member, IEEE},
        and William R. Norris \orcid{0000-0002-4940-4458}, \IEEEmembership{Member, IEEE}}

\begin{document}
\maketitle

\begin{abstract}
The emergence of tethered drones represents a major advancement in unmanned aerial vehicles (UAVs) offering solutions to key limitations faced by traditional drones. This article explores the potential of tethered drones with a particular focus on their ability to tackle issues related to battery life constraints and data latency commonly experienced by battery operated drones. Through their connection to a ground station via a tether, autonomous tethered drones provide continuous power supply and a secure direct data transmission link facilitating prolonged operational durations and real time data transfer. These attributes significantly enhance the effectiveness and dependability of drone missions in scenarios requiring extended surveillance, continuous monitoring and immediate data processing needs. Examining the advancements, operational benefits and potential future progressions associated with tethered drones, this article shows their increasing significance across various sectors and their pivotal role in pushing the boundaries of current UAV capabilities. The emergence of tethered drone technology not only addresses existing obstacles but also paves the way for new innovations within the UAV industry.
\end{abstract}

%\hspace{0.15in}


\section{Introduction}
Aerial technology has undergone notable advancements with the introduction and evolution of drone technology. Drones, once seen as a novelty have now become tools in various industries from surveillance to agriculture disaster management to entertainment. Tethered drones, which simply can be defined as drones with physical connection to a ground station through a cable called tether, are changing the usability of drones in many of these industries. The global market for tethered drones has experienced significant growth with an estimated value of US\$ 327 million in 2022 and a projected rise to approximately US\$ 1629.7 million by 2029 indicating a Compound Annual Growth Rate (CAGR) of 25.8\% \cite{vlrvc_nodate}. This increase not only reflects the increasing interest but also the expanding use of tethered drone technology across different sectors.

Tethered drones are gaining popularity for reasons that set them apart from traditional drones \cite{prior2016tethered, marques2023tethered, belmekki2022unleashing}. While conventional drones are known for their versatility and agility they face limitations related to battery life and data transmission. The finite battery capacity often results in restricted time leading to frequent recharges or replacements that disrupt continuous operations. Additionally these drones encounter issues with data latency which can hinder real time data transmission critical for applications relating to autonomous navigation.

The advent of tethered drones is changing this landscape by addressing challenges related to battery life and data latency offering improved reliability and consistent performance.
The physical connection not only ensures a constant power source but also creates a steady platform for data transfer up to 10Gb per second via fiber optics \cite{tripwireless_nodate} improving the safety and efficiency of operations.

\section{Understanding Autonomous Tethered Drones}
In industries utilizing unmanned aerial vehicles (UAVs), autonomous tethered drones mark a major technological advancement overcoming key limitations of untethered drones.

\subsection{Tethered Drones Concept and Design}
Tethered drones are UAVs linked physically to a ground station or vehicle through a cord \cite{fauzi2022development}. This cord, usually made up of power cables and data lines serves two purposes. Firstly it supplies power to the drone eliminating battery life constraints common in regular drones.
This feature allows for flight times sometimes even indefinitely which is extremely useful for continuous monitoring or surveillance missions. Additionally, the tether serves as a channel for high speed data transfer. Unlike drones that depend on wireless communication systems, which are prone to delays, signal interference and security risks, tethered drones offer a direct and reliable data connection. This guarantees data transmission without the interruptions or security vulnerabilities associated with wireless signals.

\subsection{Advancements in Technology}
The emergence of tethered drones has been supported by significant progress in drone technology and tether materials. Modern tethers can efficiently transmit power and data while being lightweight and durable to minimize any impact on the drones agility. Improvements in drone design have also led to compact and effective models capable of performing various tasks despite being physically connected by a tether. Their ability to execute tasks without human input is a result of advancements in artificial intelligence and machine learning. These technologies empower drones to navigate, adapt to conditions and even make operational decisions independently enhancing their versatility and effectiveness.

\subsection{Operational Advantages}
The benefits of using tethered drones are abundant. Their continuous power source allows for missions without the hassle of changing or recharging batteries \cite{boukoberine2019power, aktacs2019up}. This feature proves valuable in scenarios where uninterrupted aerial surveillance is crucial like during large events, prolonged surveillance tasks or ongoing monitoring of industrial areas or time-sensitive agricultural tasks.

Moreover the direct and secure data transmission function ensures that high quality data be it video feeds, images or sensor readings can be instantly and securely relayed. This capability is essential in situations requiring real time data, such as emergency responses or monitoring infrastructure in real time.

These operational advantages not only address key limitations of traditional drones but also open up new possibilities for their usage across different industries.

\section{Addressing Battery Life Limitations}
Addressing Battery Life Limitations
One of the obstacles faced by conventional drone technology is the restriction imposed by battery life. This section explores how autonomous tethered drones tackle and surpass these limitations. 

\subsection{The Challenge of Drones Reliant on Batteries}
\begin{figure}[htbp]
\centering
\includegraphics[width=0.5\textwidth]{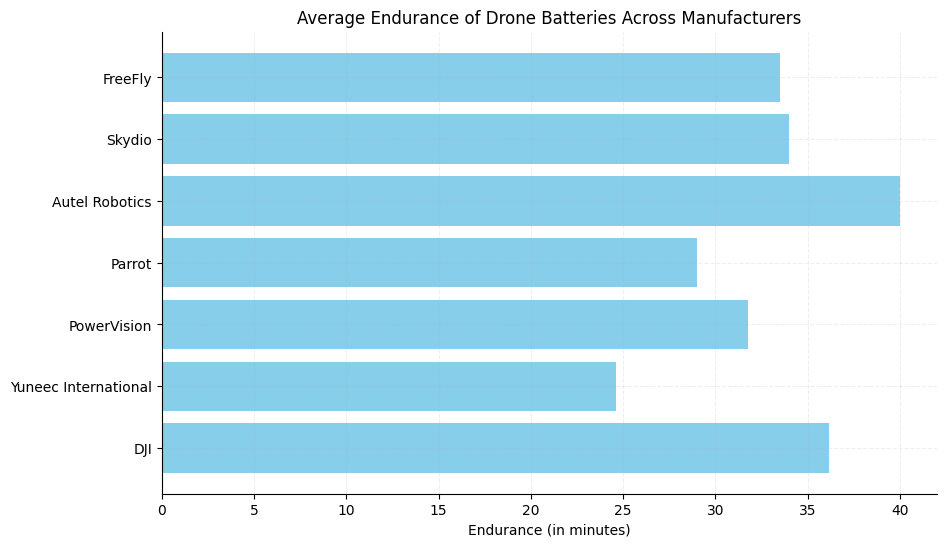}
\captionsetup{justification=centering} % Centers the caption
\caption{Average battery life of drones across manufacturers. Sources listed in Table \ref{table1}}
\label{battery_life}
\end{figure}

Relying on battery power poses a dilemma for drones. While it grants them freedom of movement and adaptability, it also imposes restrictions on their usage duration. Despite advances in technology, batteries still have an energy capacity restricting flight times to around 20 to 40 minutes before necessitating a recharge or replacement. Please see Fig.\ref{battery_life} for the average battery life of popular drone makers and Table \ref{table1} for a more detailed list. This limitation is not merely an obstacle; it can prove crucial in scenarios that demand extended aerial presence, such as ongoing surveillance, continuous data collection or lengthy inspections.

\begin{table}[htbp]

\centering
\caption{Battery life of various drone models across manufacturers.}
\label{table1}
\begin{tabular}{llrl} % four columns, no vertical lines
\toprule
\textbf{Manufacturer} & \textbf{Model} & \textbf{Battery Life (min)} & \textbf{Source} \\
\midrule
DJI & Mini 2 & 31 & \cite{dji_mini2_support} \\
& Air 2S & 31 & \cite{dji_air2s} \\
& Mini 3 & 34 & \cite{dji_mini3} \\
& Mavic 3 & 46 & \cite{dji_mavic3_support} \\
& FPV & 20 & \cite{dji_fpv} \\
& Matrice 300 & 55 & \cite{dji_matrice300_rtk} \\
\addlinespace
Yuneec International & Breeze & 12 & \cite{yuneec_battery_blog} \\
& Typhoon H & 25 & \cite{yuneec_battery_blog} \\
& Q500 4K & 25 & \cite{yuneec_battery_blog} \\
& Mantis Q & 33 & \cite{yuneec_battery_blog} \\
& H520 & 28 & \cite{yuneec_battery_blog} \\
\addlinespace
PowerVision & PowerEgg X & 34 & \cite{powervision_powereggx} \\
& Power Eye & 29.5 & \cite{powervision_powereye} \\
\addlinespace
Parrot & Anafi & 25 & \cite{parrot_anafi_usa} \\
& Bebop & 30 & \cite{droneblog_parrot_battery} \\
& Anafi Ai & 32 & \cite{parrot_anafi_ai_specs} \\
\addlinespace
Autel Robotics & Evo max & 42 & \cite{autel_evo_max_4t} \\
& Evo II & 38 & \cite{autel_evo_ii_series} \\
& Evo Lite & 40 & \cite{autel_evo_lite_series} \\
\addlinespace
Skydio & X10 & 40 & \cite{skydio_x10} \\
& X2 & 35 & \cite{skydio_x2} \\
& 2+ & 27 & \cite{skydio_2_plus_enterprise} \\
\addlinespace
FreeFly & Alta X & 30 & \cite{freefly_alta_x} \\
& Astro & 37 & \cite{freefly_astro} \\
\bottomrule
\end{tabular}
\end{table}

Furthermore, battery performance is influenced by different environmental factors. Extreme temperatures for example can notably diminish battery efficiency further curtailing flight durations. The frequent need for recharging or changing batteries leads to downtime and can interrupt the seamless execution of tasks. In operations like search and rescue missions or disaster responses or agricultural insecticide spraying where time is of the essence, these constraints can have significant repercussions.

\subsection{Tethered Drones: A Leap in Endurance}
Autonomous drones offer an innovative solution to the challenge of battery life. By drawing power from a ground based source via the tether, these drones can operate for extended periods – as long as the power supply remains uninterrupted.
This consistent power supply brings an advantage especially for tasks that require prolonged operational periods. Using this approach not only extends the flight time but also ensures reliable performance. Unlike batteries that can lose power over time the constant and dependable power supply from the tether enables the drone to function efficiently throughout its mission. 

\subsection{Enhanced Operational Efficiency}
The longer operational duration of drones enhances their overall efficiency. By eliminating the limitation of battery life these drones can be deployed for missions without frequent interruptions for recharging. This continuous operation allows for thorough data collection, uninterrupted surveillance and prolonged inspection tasks without interruptions or downtime associated with changing batteries.

Moreover this enhanced efficiency results in cost savings in the long term. By reducing the need for multiple drones to cover missions and minimizing downtime, operational costs are lowered along with resource utilization.

\section{Overcoming Data Latency}
Apart from tackling battery life issues, autonomous tethered drones also play a role in addressing the problem of data latency, which significantly impacts the efficiency of drone operations. This section explores how these advanced drones offer a solution to data transmission delays that often hinder drones.

\subsection{Data Latency in Conventional Drones}
Data latency refers to the time gap between when the drone captures data through onboard sensors and when it is sent to the operator or processing center. In drones that rely on wireless communication systems, various factors contribute to data latency, including signal interference, bandwidth constraints and the distance between the drone and the receiver. In situations like emergency responses or real time surveillance even a slight delay can hinder operation effectiveness potentially leading to risks or missed opportunities.

Moreover, wireless transmission is vulnerable to security threats. The transmitted data can be breached, posing risks to sensitive information confidentiality and integrity. This poses concerns in fields such as law enforcement, border monitoring and corporate espionage.

\subsection{The Tethered Advantage}
Autonomous tethered drones present a solution to these challenges. The physical tether establishes a secure link, for data transmission. 

This direct physical link ensures data transfer effectively even up to 10Gb/s for fiber optics cables removing the delay linked with wireless communications. The immediacy of this data transfer is crucial in situations where real time analysis and prompt decision making are vital.

Furthermore, the tethered system offers a pathway for data transmission. The physical connection is less susceptible to interception or hacking when compared to wireless signals thus significantly boosting data security. This is especially important in activities that demand levels of privacy like military operations or monitoring critical infrastructure.

\subsection{Enhanced Data Transmission Capabilities}
The enhanced data transfer capabilities of drones go beyond just reducing delays. They also enable the transmission of larger amounts of data and support higher quality data streams. For example, the stable connection facilitates the transfer of high definition video feeds which are essential for detailed surveillance and inspection tasks.

Additionally, the reliability of the connection ensures consistent performance even in challenging environments where wireless communications could be compromised such as heavily populated urban areas or remote locations with limited connectivity.

Autonomous tethered drones effectively address the data latency challenges often encountered in drone operations. By enabling time, secure and high capacity data transmission, they facilitate more efficient and productive operations, across various applications. This advancement in technology not only improves drone capabilities but also broadens its potential applications making it a valuable tool in our increasingly data centric world.

\section{Challenges and Future Prospects}
While autonomous tethered drones offer advantages over traditional models, they do come with their own set of challenges. This section looks into these limitations and the future possibilities of this technology outlining both the challenges and the potential advancements on the horizon.

\subsection{Challenges Facing Tethered Drones}
\begin{itemize}

\item Mobility Limitations: One key challenge is the mobility caused by the physical tether. Although it ensures power and secure data transmission, it constrains the drones operational range. Navigating environments or covering extensive areas can be more demanding for tethered drones compared to their untethered counterparts. Future research can consider integrating the tethering station on a moving platform to enhance mobility and coverage.

\item Tether Management: The handling of the tether itself presents challenges. Ensuring that it remains untangled, undamaged and does not hinder operations requires planning and at times additional equipment. Future research in this area can explore advanced and model-based control systems to ensure the tether is always taut and not too taut to affect the drone navigation.

\item Deployment Time: Tethered drones may require setup times more than battery operated drones. The necessity to set up the tether system could potentially delay readiness, in time sensitive scenarios. Research in this area could explore automating the deployment system with minimal human interactions. 

\item Weather and Environmental Impact: Although tethered drones are more resistant to conditions compared to battery powered drones, they can still be affected by extreme weather impacting both their performance and the durability of the tether. For example, the tether might be affected by a gause of wind thereby affecting the stability of the drone. Research in this area should explore incorporating disturbances like wind and drag into the dynamics of both the drone model and the tether model.
\end{itemize}

\subsection{Future Prospects and Technological Developments}

Despite these obstacles the future of tethered drones appears promising, with various avenues for technological progress;

\begin{itemize}

\item  Advanced Tether Materials and Designs: Research focusing on developing more flexible, sturdy and lightweight tether materials could improve the agility and range of tethered drones. Innovations in tether design may involve mechanisms for deployment and automated management systems for tethers.

\item  Improved Autonomy and AI Integration: Progress in intelligence and self-guided navigation will further boost the capabilities of these drones making them more effective and adaptable in intricate operational settings.

\item  Expanded Applications: With advancements, new uses for tethered drones are likely to emerge. This could involve utilizing them in communication networks, traffic control systems and extensive environmental monitoring efforts.

\item  Integration with Other Technologies:  The potential integration of tethered drones with emerging technologies, like 5G networks, IoT devices and advanced sensors could introduce fresh opportunities while enhancing their operational efficiency. 

\end{itemize}

\section{Conclusion}
This article explored the emerging technology of tethered drones and their abilities to revolutionize the functionalities and uses of unmanned aerial vehicles. The examination has shed light on how these drones tackle the constraints of traditional battery operated drones especially concerning operational endurance and data transmission reliability. With their tethered configuration enabling power supply and secure real time data transfer, new avenues for extended and effective drone operations are unlocked.
Nevertheless the path of tethered drones is not without its hurdles. Challenges like mobility, tether management issues and susceptibility to environmental factors pose barriers that must be overcome. Despite these potential limitations, the future looks promising for tethered drones as advancements in materials, autonomy capabilities and integration with state of the art technologies pave the way for broader applications and enhanced functionalities.
The importance of tethered drones, in overcoming the inherent limitations of standard drones cannot be emphasized enough. With the rising need for secure aerial operations in different industries, tethered drones play a vital role. It is essential for both scientific and industrial sectors to continue exploring and enhancing this technology, expanding its capabilities. Embracing and improving tethered drone technology not only enhances UAV functions but also paves the way for new opportunities in a society increasingly dependent on autonomous systems.

% References section
\bibliographystyle{ieee} 
\bibliography{references} 

%\vspace{11pt}

\begin{IEEEbiography}[{\includegraphics[width=1in,height=1.25in,clip,keepaspectratio]{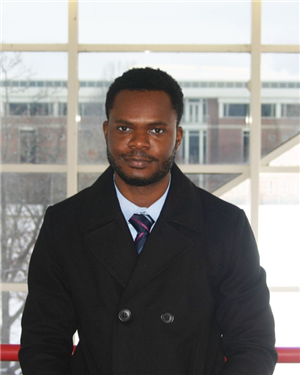}}]{SAMUEL O. FOLORUNSHO is a Graduate member of IEEE and
a PhD student conducting research at the Autonomous and
Unmanned Vehicle Systems Lab (AUVSL) and the Center for Autonomous
Construction and Manufacturing at Scale (CACMS) in the Department of
Systems Engineering at the University of Illinois, Urbana-Champaign (UIUC) under the advisorship of Prof. William R. Norris. His research is focused on control systems, computer vision and robotics - and the intersection of those for
safety-critical systems in industrial and agricultural applications.
He earned his M.S. from UIUC in 2023 and his B.S. in 2017 at
the University of Ilorin, Nigeria both in Agricultural and Biological
Systems Engineering. He has 3 years of working experience in management consulting.}
\end{IEEEbiography}

\vspace{22pt}

\begin{IEEEbiography}[{\includegraphics[width=1in,height=1.25in,clip,keepaspectratio]{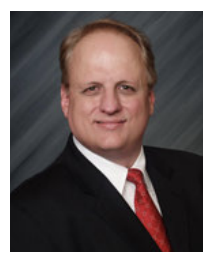}}]{WILLIAM R. NORRIS (Member, IEEE) received
the B.S., M.S., and Ph.D. degrees in systems engineering from the University of Illinois at Urbana–
Champaign, in 1996, 1997, and 2001, respectively,
and the M.B.A. degree from the Fuqua School of
Business, Duke University, in 2007. He has over
23 years of industry experience with autonomous
systems. He is currently a Clinical Associate Professor with the Industrial and Enterprise Systems
Engineering Department, University of Illinois at
Urbana–Champaign, the Director of the Autonomous and Unmanned Vehicle
System Laboratory (AUVSL), as well as the Founding Director of the Center
for Autonomous Construction and Manufacturing at Scale (CACMS).}
\end{IEEEbiography}

\end{document}